\documentclass{article}

\usepackage{amssymb,latexsym}

\usepackage[pdftex]{graphicx}

\usepackage{hyperref}
\usepackage{xcolor}
\hypersetup{
    colorlinks,
    linkcolor={red!50!black},
    citecolor={blue!50!black},
    urlcolor={blue!80!black}
}

\begin{document}

\pagestyle{plain}

\newtheorem{theorem}{Theorem}[section]

\newtheorem{proposition}[theorem]{Proposition}

\newtheorem{lemma}[theorem]{Lemma}

\newtheorem{corollary}[theorem]{Corollary}

\newtheorem{definition}[theorem]{Definition}

\newtheorem{remark}[theorem]{Remark}

\newtheorem{exempl}{Example}[section]

\newenvironment{example}{\begin{exempl}  \em}{\hfill $\square$

\end{exempl}}  \vspace{.5cm}

\renewcommand{\contentsname}{ }

\title{Artificial life properties of directed interaction combinators vs. chemlambda}

\author{Marius Buliga \\ 
\\
Institute of Mathematics, Romanian Academy \\
P.O. BOX 1-764, RO 014700\\
Bucure\c sti, Romania\\ 
{\footnotesize Marius.Buliga@imar.ro , mbuliga@protonmail.ch}}  \vspace{.5cm}

\date{12.05.2020}

\maketitle

\begin{abstract}
We provide  a framework for experimentation \href{https://mbuliga.github.io/quinegraphs/ic-vs-chem.html#icvschem}{at this link} with two artificial chemistries: directed interaction combinators (dirIC, defined in section \ref{chemistries}) and chemlambda \cite{buligachemlambda}. We are interested if these chemistries allow for artificial life behaviour: replication, metabolism and death.

The main conclusion of these experiments is that graph rewrites systems which allow conflicting rewrites are better than those which don't, as concerns their artificial life properties. This is in contradiction with the search for good graph rewrite systems for decentralized computing, where non-conflicting graph rewrite systems are historically preferred.

This continues the artificial chemistry experiments with chemlambda, lambda calculus or interaction combinators, available from the entry page \href{https://chemlambda.github.io/index.html}{chemlambda.github.io} \cite{entry} and described in arXiv:2003.14332.
\end{abstract}

\section{Molecular computers as graph rewriting local machines}

We define  a molecular computer \cite{molecular} as one molecule which transforms, by random chemical reactions mediated by a collection of enzymes, into a predictable other molecule, such that the output molecule can be conceived as the result of a computation encoded in the initial molecule.

The goal is to be able to make real life molecular computers. We use artificial chemistry to understand how to do this.

Molecules are port graphs in the sense of Bawden \cite{bawden1}, \cite{bawden2} with nodes decorated with a finite set of node types. The chemical reactions are double push-out graph rewrites \cite{dpo} mediated by enzymes:

\vspace{.5cm}

\begin{center}
(LHS) + (TOKEN\_1) + (ENZYME) = (RHS) + (TOKEN\_2) + (ENZYME)
\end{center}

\vspace{.5cm}
$\left.\right.$ \\
where (LHS) and (RHS) are the left hand side and right hand side pattern of the graph rewrite and (TOKEN\_1) and (TOKEN\_2) are molecules which serve to make the reaction to conserve the nodes and links. 
Here we do not represent these tokens, nor the enzymes in the simulations.

The graph rewrites belong to a universal family, whose members appear everywhere in mathematics and physics and computer science. All of them are of the kind encountered in Interaction Combinators \cite{lafont-comb}, but they may differ by the choice of node types or port types.

\paragraph{1st hypothesis:} these rewrites can also be used to make real molecular computers.

\paragraph{2nd hypothesis:} the enzymes create a graph rewriting algorithm which can be implemented by local machines.

Here a local machine is a defined as a Turing machine with two tapes: the IO tape and the work tape. Molecules are encoded as mol patterns, which are vectors of mol nodes, where each mol node is a vector of node types and tags of edges connected to the node ports. The tapes of the machine have cells, each cell of a tape can contain a node type, a port type, an edge tag or a finite set of special characters.

The machine can read from on the IO tape a cell, write a whole mol node at the end of the mol pattern, or delete a whole mol node. For each read/write or delete from/to the IO tape the machine writes the same on the work tape. The machine can read from the work tape and it can write only on the blank cells of the work tape. The machine can either halt (by writing a halt symbol on the work tape) or it can delete the whole work tape and then the IO head jumps to a random mol node beginning.

The constraints are:
\begin{enumerate}
\item[-] the machine can perform any graph rewrite on any molecule
\item[-] at each moment the IO tape contains only a well written mol pattern
\item[-] the work tape has finite length.
\end{enumerate}

A graph rewriting algorithm which can be implemented by a local machine is by definition local in space, in the sense that the graph rewrites can be performed by using a local machine (therefore by a TM with a finite work tape), and local in time, in the sense that the algorithm can retain only a limited in time portion of the history of the graph evolution.

In these simulations we use a random rewrite algorithm and we also randomize as many times as necessary the data (mol file, list of possible rewrites,...) in order to destroy all information which cannot be processed by a local machine. As an example, the list of nodes in the mol file is ordered, but we have to avoid this information and instead to jump randomly to a node or explore from a node only a limited portion of the graph. The only exception we make is to be able to use an age of the node or of the link of the graphs, for the purpose of searching for quine graphs (see below about the use of the "change" button for the "older is first" option). We could, instead use a local algorithm based on age, where there is a upper age parameter such that the age of a node or link which is bigger than the upper age trigger an age reset (and also it may trigger a rewrite, like deletion of the old node or link). However, in these experiments we don't go in this direction.

\section{Chemlambda and Directed IC artificial chemistries}
\label{chemistries}

Here we compare two graph rewrites systems and their associated artificial chemistries. We use the term artificial chemistry as it applies to chemical reactions of individual molecules, that is in the sense of asynchronous graph rewrite automata \cite{tomita}. In \cite{banzhaf}, \cite{dittrich},  artificial chemistries are relevant for populations of molecules.  

The first is chemlambda \cite{buligachemlambda}.  Chemlambda appeared first as the "chemical concrete machine" \cite{buligachem}, a name inspired by the chemical abstract machine \cite{cham} of Berry and Boudol.  In collaboration with Kauffman \cite{chemlambda1} appeared a non-local version of chemlambda graph rewrite system, but without any algorithm of application. In that paper we were interested in the relations with knot theory graph rewrites. Kauffman wrote previously about knot automata \cite{kauffman1} and knot logic \cite{kauffman2}. 
Chemlambda as an artificial chemistry, with purely local graph rewrites is described at length in \cite{buligachemlambda} and available for experimentation in \cite{entry}. 

From a historical perspective, Fontana and Buss \cite{fontana1}, \cite{fontana2}, build their Alchemy (algorithmic chemistry) based on  populations of lambda calculus terms in irreducible form. As chemical reactions  they take lambda calculus application, followed by reduction to the normal form. Chemlambda is a different artificial chemistry, where individual molecules (chemlambda graphs) interact with rewrite enzymes. Chemlambda can be used for lambda calculus reductions, see \cite{lambda2mol}.

 The second is Directed Interaction Combinators (dirIC), which has a double origin. In Lafont' Interaction Combinators \cite{lafont-comb}, section 3.4, is also described a directed version. We used this proposal to provide a parsing from IC nodes to chemlambda nodes:   

\vspace{.5cm}
 
\centerline{     \includegraphics[width=  70.8mm]{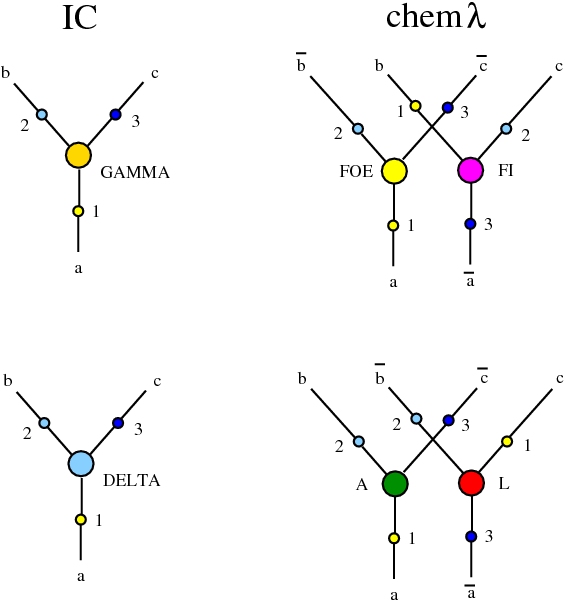}}

\vspace{.5cm}
$\left.\right.$ \\
which works well if we renounce at the chemlambda rewrites A-FO and A-FOE 

\vspace{.5cm}

\centerline{     \includegraphics[width=  80.8mm]{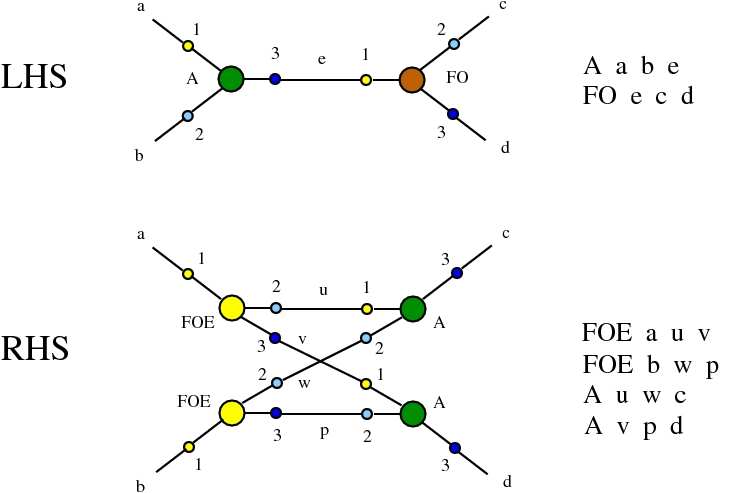}}

\vspace{.5cm}

\centerline{\includegraphics[width=  80.8mm]{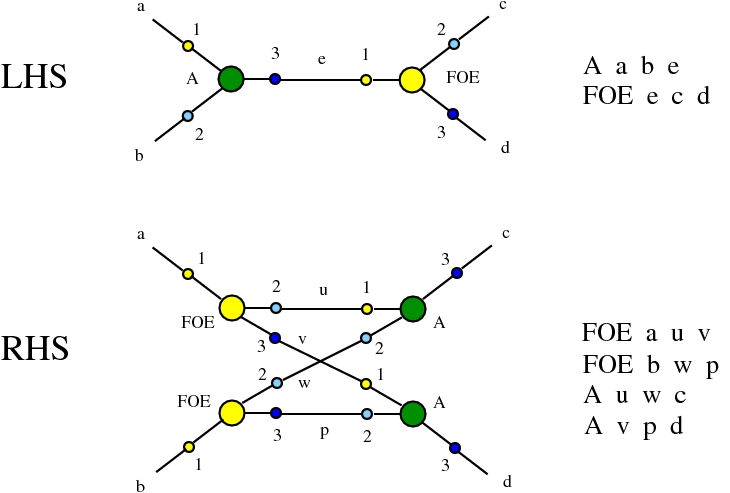}}

\vspace{.5cm}
$\left.\right.$ \\
and we replace them by a rewrite FI-A.

\vspace{.5cm}

\centerline{     \includegraphics[width=  80.8mm]{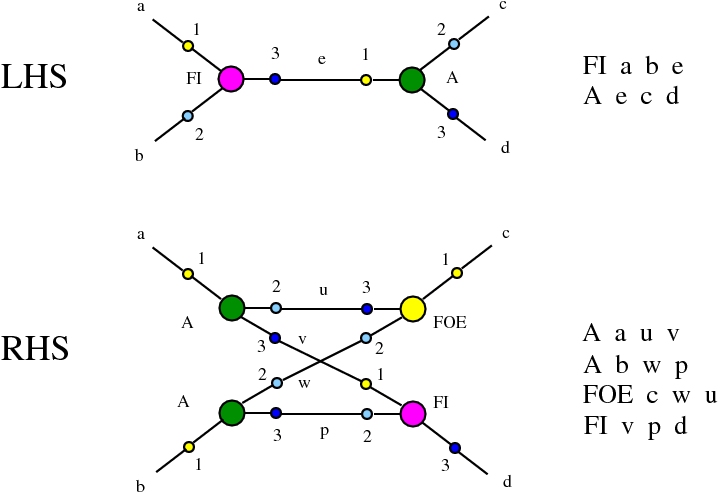}}

\vspace{.5cm}

Some termination rewrites are modified as well. More precisely, these rewrites are described in the file \href{https://github.com/mbuliga/quinegraphs/blob/master/js/chemistry.js}{chemistry.js}. Chemlambda rewrites are obtained by concatenation of CHEMLAMBDABARE with CHEMLAMBDAEND and dirIC rewrites are obtained from concatenation of CHEMLAMBDABARE with DICMOD. The resulting graph rewrite system is very close to Asperti' BOHM machine graph rewrite system \cite{bohm}.

For the purposes of this work, the main difference between these two graph rewrite systems is that chemlambda has conflicting rewrite (patterns) and dirIC does not have such patterns. The cause of this difference is that some chemlambda nodes have two active ports while all dirIC nodes (as well as the original IC nodes) have only one active port.

\section{Chemlambda quines vs Directed IC quines}

\paragraph{Problem:} find those graph rewrite systems which admit artificial life, i.e. graphs (molecules) with the properties
\begin{enumerate}
\item[-] they have metabolism (graph quines)
\item[-] they can duplicate
\item[-] they can die.
\end{enumerate}

Here a graph quine is a molecule which admits a periodic evolution under graph rewrites, provided the local machine is constrained to rewrite older graph rewrites patterns first and to choose one among the possible strategies: whenever possible to prefer the rewrites which grow the number of nodes ("GROW") vs those which reduce the numebr of nodes ("SLIM") or inversely.

Once we have a graph quine, we are interested if such a molecule can duplicate or die if the local machine is not constrained by the age of the rewrites patterns or by the choice of a strategy.

Death of a quine graph is understood as the evolution to a state where no more graph rewrites are possible.

We see in these experiments that the choice of the graph rewrite system is important if we want to have graphs with these artificial life properties. 

We can experiment with 10 chemlambda quines and with 5 Interaction Combinators quines, on the same ground. There are comments relevant for each quine, which appear when the user chooses a quine from the menu. We use the same list of quines as those initially presented in \cite{ice}.

There are examples which show that a chemlambda quine may be as big as we wish (the "ouroboros quine"), that it can duplicate (the 10\_nodes quine), that there exist non-connected quines (the 16\_quine\_A\_L\_FI\_FO) which are not made by connected components which are quines.

IC quines are given as well, starting with Lafont' quine, which appears in the foundational article about interaction combinators, and continuing with IC quines discovered from random search in a big family of IC graphs.

There are also options to search for new quines among random graphs.

For the IC quines we can transform them into dirIC graphs, which have nodes compatible with chemlambda. For any of these quines we can see their evolution under chemlambda or under dirIC.

In order to verify if a graph is a quine we need to do the following, as described in \cite{quinegraph}:
\begin{enumerate}
\item[-] click on the first "change" button on the left to see the message "older is first".
\item[-] choose among the two extremal strategies "GROW" or "SLIM" by moving the rewrites weights slider. For chemlambda or IC quines the choice is "GROW".
\item[-] choose among the chemistries dirIC or chemlambda by using the second "change" buton.
\end{enumerate}
If the evolution of the graph is periodic then we have a quine.

These quines have a metabolism, in the following sense. We use the first "change" button in order to have "random choices" and we move the rewrites weights slider to the middle (or we can position it as we like in different experiments, or even during the evolution of the graph). We are now in the random rewrites algorithm. A graph (in particular a quine graph) exhibits metabolism if it has an approximately periodic evolution for an ammount of time, followed perhaps by death. Death is lack of rewrites available.

We see that chemlambda quines as well as dirIC quines exhibit metabolism, but only chemlambda quines may die. The lack of conflicting rewrite patterns induces a unique final state of a graph, when it exists, which is a desired property in distributed computing. The same unicity is not desired in artificial life.

We achieved replication in chemlambda but not in dirIC, probably due to a lack of enough experiments.

A curious phenomenon is that some IC quines do not translate to dirIC quines. This is due to the fact that the translation from IC to dirIC works as expected for graph which have a final state where there is no possible reduction left. In the case of an IC quine, the reduction does not terminate, it is periodic. When translated to a dirIC graph we know that the reduction can't terminate, but there is no guarantee that the reduction is periodic. We only know that if we synchronize the rewrites (so as to respect the fact that a rewrite in IC is two rewrites in dirIC) then we shall have a periodic evolution.

Some chemlambda quines are dirIC quines as well and inversely. But more times (as shown by experiments) a chemlambda quine dies fast in the process of quine checking (i.e. "GROW" strategy and "older is first"). Namely this happens for chemlambda quines which can die (attain a final state) under the rewrites from CHEMLAMBDABARE.

Finally, if we want to explore the computational properties of these chemistries, see \cite{lambda2mol}, which is a parser and reducer which has the option to choose the chemistry for reduction of lambda calculus terms. Under this simple rewriting algorithm, chemlambda is better than dirIC, but nevertheless the dirIC has many cases where it reduces lambda terms as well.

\end{document}